\documentclass[journal]{IEEEtran}
\usepackage{times}
\usepackage{epsfig}
\usepackage{graphicx}
\usepackage{array}
\usepackage{algorithmic}
\usepackage{amsmath}
\usepackage{amssymb}
\usepackage{tabularx}
\usepackage{multirow}
\usepackage{booktabs}
\usepackage{arydshln}
\usepackage[dvipsnames]{xcolor}
\usepackage{fontawesome}
\usepackage{caption}
\usepackage{subcaption}
\usepackage{enumitem}
\usepackage{pifont}
\usepackage{xspace}
\usepackage{tabularx}
\usepackage{multirow}
\usepackage{caption}
\usepackage{booktabs}
\usepackage{cite}
\usepackage[pagebackref=true,breaklinks=true,letterpaper=true,colorlinks,bookmarks=false]{hyperref}
\usepackage[capitalize]{cleveref}
\crefname{section}{Sec.}{Secs.}
\Crefname{section}{Section}{Sections}
\Crefname{table}{Table}{Tables}
\crefname{table}{Tab.}{Tabs.}

\makeatletter
\DeclareRobustCommand\onedot{\futurelet\@let@token\@onedot}
\def\@onedot{\ifx\@let@token.\else.\null\fi\xspace}
\def\eg{\emph{e.g}\onedot} 
\def\ie{\emph{i.e}\onedot}

\makeatother
\newcommand{\cmark}{\ding{51}}%
\definecolor{ourgreen}{RGB}{56,87, 35}
\definecolor{ourorange}{RGB}{184,97, 35}
 
\begin{document}

\title{3DoF Localization from a Single Image and an Object Map: the Flatlandia Problem and Dataset 
\thanks{This project has received funding from the European Union's Horizon 2020 research and innovation programme under grant agreement No 870743 and the project Future Artificial Intelligence Research (FAIR) – PNRR MUR Cod. PE0000013 - CUP: E63C22001940006.}}

\author{Matteo~Toso, 
        Matteo~Taiana, 
        Stuart~James,
        and~Alessio~{Del~Bue} 
}

\maketitle

\begin{abstract}
Efficient visual localization is crucial to many applications, such as large-scale deployment of autonomous agents and augmented reality. Traditional visual localization, while achieving remarkable accuracy, relies on extensive 3D models of the scene or large collections of geolocalized images, which are often inefficient to store and to scale to novel environments. In contrast, humans orient themselves using very abstract 2D maps, using the location of clearly identifiable landmarks. Drawing on this and on the success of recent works that explored localization on 2D abstract maps, we propose Flatlandia, a novel visual localization challenge. With Flatlandia, we investigate whether it is possible to localize a visual query by comparing the layout of its common objects detected against the known spatial layout of objects in the map. We formalize the challenge as two tasks at different levels of accuracy to investigate the problem and its possible limitations; for each, we propose initial baseline models and compare them against state-of-the-art 6DoF and 3DoF methods. 
Code and dataset are publicly available at \url{github.com/IIT-PAVIS/Flatlandia}.
\end{abstract}


\IEEEpeerreviewmaketitle

\section{Introduction}\label{sec:intro}

\IEEEPARstart{F}{rom} the Babylonian's stone-carved Imago Mundi to the GPS-based Google Maps app used by millions of users every day, humans have relied on bi-dimensional, abstract representations to describe the world around them. These provide a simplified model of the environment based on the location of clearly identifiable landmarks; this is consistent with how humans think about their location, relying on fixed visual landmarks for self-localizing~\cite{tiwari2012voice}.

In this spirit, 3DoF visual localization~\cite{noe2020eccv, Vojir_2020_ACCV, sarlin2023orienternet} regresses the \emph{3DoF pose} (\ie longitude, latitude and orientation) of a camera on an abstract \emph{reference map} of the scene, given a visual query as input. 
Such maps are more efficient to generate, store and update than the scenes representation typically used by 6DoF methods~\cite{kendall2015posenet, sarlin2019coarse, arandjelovic2013all}. Most approaches represent the scene as a 3D model or a collection of reference images with known pose, both of which require storing hundreds of MB even for small scenes~\cite{Sattler_2018_CVPR}; abstract maps, in contrast, are typically composed of a collection of polygons, lines and points, representing building footprints, roads and common objects, respectively. 
This makes 3DoF localization a good candidate for real-world applications like large-scale deployment of autonomous agents or Augmented Reality (AR).

In this work, we propose \emph{Flatlandia}, a novel challenge for 3DoF localization from object detections. Given a query image, we want to estimate its 3DoF location by comparing the estimated layout of the objects detected in the image (\emph{local map}) against the known GPS locations of objects in the scene (\emph{reference map}). Describing the problem as sets of 2D detections (\emph{query}) and point-like objects on a 2D plane (\emph{reference}) introduces privacy preserving properties and significantly reduces  storage requirements. For example, on average the scenes used in this paper require $50$~kB to store the object map, but $76$~MB and $271$~MB to store the scene's 3D point cloud and reference images, respectively.

\begin{figure*}[th!]
        \centering
        \includegraphics[trim={0 2cm 0 2cm}, clip, width=\linewidth] {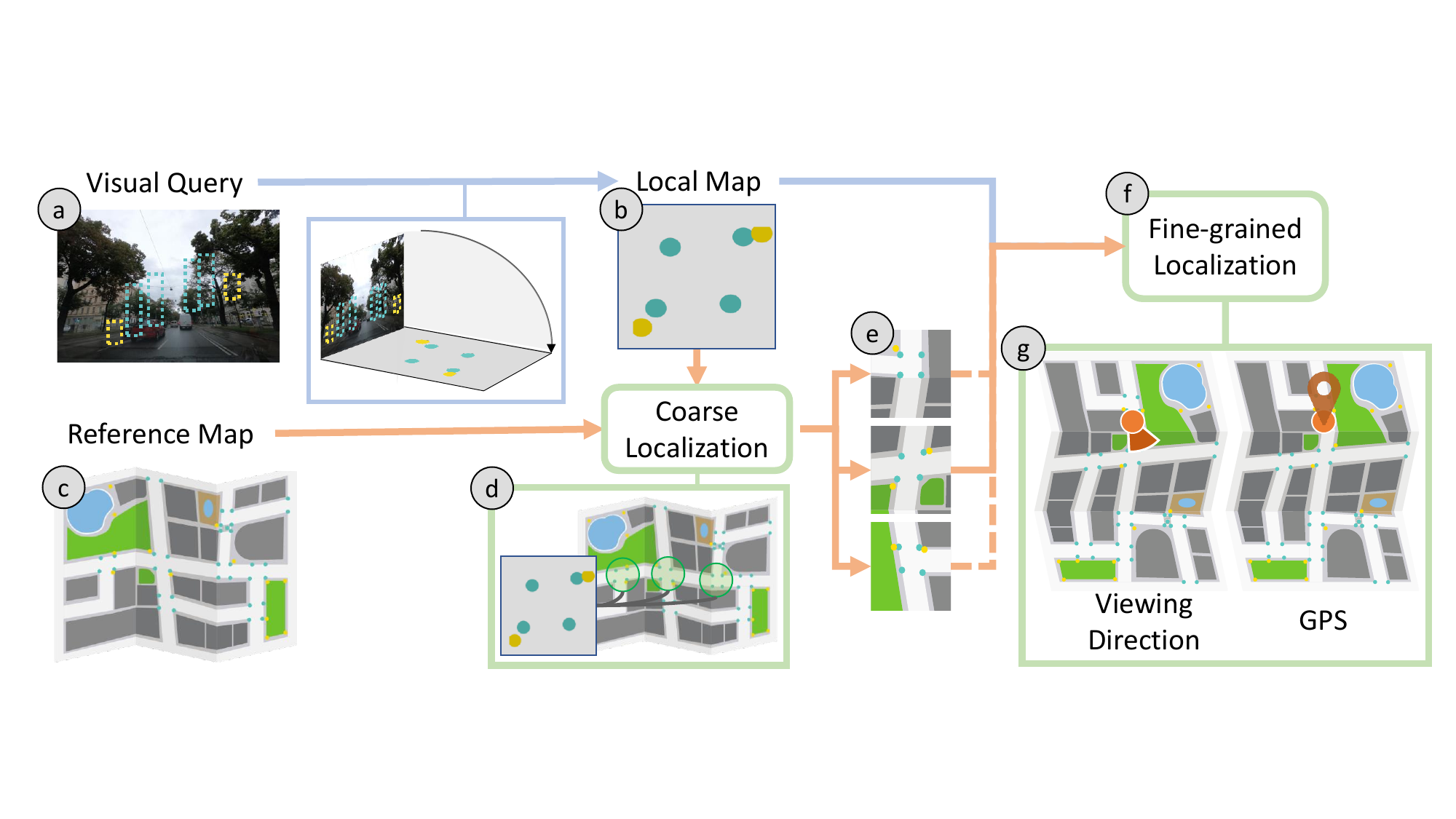}
    \caption{\emph{The Flatlandia 3DoF Localization Problem}: Given a visual query with object detections (a), we estimate the layout of the observed objects with respect to the camera, representing it as 2D points on a bird's-eye-view \emph{local map} (b). We then compare this map against a \emph{reference map} annotated with the known location of objects in the scene (c): first a \emph{Coarse Localization} module (d) generates proposals or \emph{reference} map regions whose object distributions are compatible with the query (e). Then, a \emph{Fine-grained Localization} module (f) compares each proposed region against the \emph{local} map, to estimate the \emph{GPS} location and the \emph{viewing direction} of the query in the \emph{reference map} coordinate system (g).
    }\label{fig:mainfig}
\end{figure*}

We propose to divide the problem in two tasks, as shown in Fig.~\ref{fig:mainfig}. First, \emph{coarse map localization} (Sec.~\ref{sec:coarseloc}) explores limitations due to repeated patterns, attempting to reliably retrieve a region of the reference map containing the same layout of the query objects; then, \emph{fine-grained localization} (Sec.~\ref{sec:fineloc}) combines the output of the first task (\emph{region proposal}) and the query to regress the 3DoF camera pose, investigating limitations due to representing objects as single points and possible mismatches. For both tasks, we provide example baseline approaches and their evaluation (Sec.~\ref{sec:experiments}).

One additional obstacle to object-based 3DoF localization is that existing datasets are aimed towards different applications. The traditional datasets used for 6DoF do not contain large enough maps or object detections and their 3D location; on the other hand, existing 3DoF methods use OpenStreetMap\footnote{OpenStreetMap: \href{https://www.openstreetmap.org}{www.openstreetmap.org}} (OSM) coupled with geolocalized images from other datasets, which do not provide Ground Truth (GT) matches between images and object annotations on the map. Moreover, OSM object annotations are provided by users, making location accuracy and annotation abundance inconsistent from location to location. To provide a benchmarking environment for Flatlandia, we develop the Flatlandia dataset\footnote{Flatlandia Dataset: \href{https://www.github.com/IIT-PAVIS/Flatlandia}{www.github.com/IIT-PAVIS/Flatlandia}} (Sec.~\ref{sec:dataset}), composed of 20 large-scale ($0.025$ km$^2$ each) \emph{reference maps} from five European cities, with associated visual queries and \emph{local maps}.

The contributions of this paper are, therefore, four-fold: \emph{i)} we  define the Flatlandia problem, a novel visual localization challenge aimed to foster the development of efficient and privacy-preserving methods; \emph{ii)} we formalize the challenge as two tasks, \emph{coarse} and \emph{fine-grained} localization, to investigate different limitations of the proposed problem; \emph{iii)} we release a novel dataset to benchmark object-based 3DoF localization; and \emph{iv)} we  exemplify the storage and computational advantages of Flatlandia by evaluating seven different baseline methods on the two aforementioned tasks, and compare them against popular 3DoF and 6DoF methods on our dataset.

\section{Related Work}\label{sec:relatedwork}
Visual Localization is a well-studied problem in Computer Vision. This section provides an overview of Visual Localization (Sec.~\ref{sec:related_visloc}), and of relevant works on object-based localization (Sec.~\ref{sec:related_objectloc}), 3DoF localization on 2D maps (Sec.~\ref{sec:related_maploc}), and datasets for visual localization (Sec.~\ref{sec:related_dataset}). See~\cite{9336674} for a full review of classical 6DoF visual localization.

\subsection{Visual Localization}~\label{sec:related_visloc}

One of the most common visual localization techniques is~\emph{image retrieval}~\cite{humenberger2022investigating,Hausler_2021
}, which compares the \emph{visual query} (\ie image to localize) against an existing database of images whose 6DoF camera pose is known. By encoding both query and reference images visual information in a global embedding (\eg DenseVLAD~\cite{7298790}, NetVLAD~\cite{7937898}), the visual query pose can be approximated as the pose of the closest image in embedding space, or as an interpolation of the closest $N$ images~\cite{torii2019large
}. The density of reference images in the scene limits the accuracy of these methods, however, this initial estimate can be further refined by regressing the relative camera pose between query and retrieved image via Convolutional Neural Networks (CNNs)~\cite{LMKK2017ICCVW} or by regressing and combining their essential matrices~\cite{zhou2020learn}. Other works exploit an explicit model of the scene, \eg a sparse SfM reconstruction derived from the available reference images~\cite{
Dusmanu_2019_CVPR, Cheng_2019_ICCV, sarlin2019coarse}. This can be used in combination with image retrieval, \eg by estimating the relative query-to-retrieved camera poses using perspective-n-point~\cite{sarlin2019coarse, sarlin2021back} (PnP), or to directly establish matches between 2D detections on the visual query and 3D points of the model exploiting depth images~\cite{shotton2013scene} or re-identifying pre-defined anchor points in the scene~\cite{anchorpoint}. 
Alternatively, models like PoseNet~\cite{kendall2015posenet} implicitly learn to model a scene, exploiting advances in CNNs to directly regress the camera pose from images; subsequent work also exploits geometry-aware loss functions (Geo-PoseNet~\cite{Kendall_2017_CVPR}), or account for the uncertainty of the predicted pose (Bayes-PoseNet~\cite{Kendall2016ModellingUI}).

These approaches require storing a large amount of data, which can limit their scalability. For image retrieval, even a small area can require a large amount of reference images ($3.1$~GB for $0.5$~km radius~\cite{Sattler_2018_CVPR}); models like PoseNet~\cite{kendall2015posenet} require only to store the neural network's weights (approximately $150$~MB), but a model has to be trained for each scene. Moreover, image retrieval relies on accurate keypoint matching; in urban environments, repeated patterns, dynamic elements and environmental changes between data acquisition and deployment can cause mismatches and misdetections even for state-of-the-art methods like SuperGlue~\cite{sarlin20superglue} and LightGlue~\cite{lindenberger2023lightglue}.

\subsection{Object-based Localization}~\label{sec:related_objectloc}
Some works address this by exploiting semantic information.
This includes representing the scene as clouds of 3D objects, parameterized as ellipsoids~\cite{zinshal02975379}, and estimating the camera pose from perspective geometry~\cite{crocco2016structure, 8440105}; or localizing cameras by modeling and re-identifying local buildings~\cite{Xue_2022_CVPR}. Semantic information can also provide correspondences across images~\cite{cathrin}, which can be used to guide local feature matching, making it more robust~\cite{benbihi2022}.

\subsection{Localization on 2D Maps}~\label{sec:related_maploc}
Other approaches model the scene through 2D reference maps, which are cheaper to store than extensive 3D models; the localization problem is then framed as the estimation of a 3DoF pose (\ie location and orientation) on a 2D map. Existing works~\cite{9635972,noe2020eccv,Vojir_2020_ACCV, sarlin2023orienternet} use reference maps from OSM, which provide vector tiles with objects (polygonal areas, multi-segment lines, or single points) representing various urban elements (\eg building outlines, roads, and objects, respectively). Such map tiles are then used to create embeddings which are compared against the embeddings of the visual query to estimate a 3DoF pose on the map; this is facilitated by the use of panoramic ($360^\circ$) images~\cite{9635972, noe2020eccv}. Alternatively, localization can be driven by the layout of buildings and approached as an instance re-identification problem by comparing embeddings representative of the distribution of buildings around a point on the map against panoramic images coupled with depth estimation~\cite{Vojir_2020_ACCV}. Proposed more recently, OrienterNet~\cite{sarlin2023orienternet} regresses the 3DoF pose from a visual query by comparing bird's-eye view neural maps generated via graph neural networks from a query and from nearby OSM data. While this method can work on single image inputs, it achieves its best performance when localizing a sequence of images, like done by~\cite{sarlin2023orienternet}. In~\cite{sarlin2023orienternet}, using sequences instead of single images increases the fraction of queries localized within $1$~m of the GT location from $54\%$ to $80\%$.

In contrast, we explore the possibility of using only the known location of objects in the scene and a single visual query, constraining the amount of input data required. This is a significantly harder task, which requires the development of novel techniques both to represent the query scenes and to match query and reference maps. OrienterNet is the work that most closely matches Flatlandia's objective, as it takes as input only one image and a coarse location to query OSM. While their method uses additional multi-modal information respect to our concise object maps, it is currently the best available candidate for comparison on the proposed dataset.

\begin{table}[b!]
    \scriptsize
     \begin{tabularx}{\linewidth}{l@{\hskip 0.0 in} c@{\hskip 0.0 in} c@{\hskip 0.1 in} c@{\hskip 0.1 in} c@{\hskip 0.1 in} c@{\hskip 0.1 in} c@{\hskip 0.1 in} c@{\hskip 0.1 in} c@{\hskip 0.1 in} r}
     \toprule
    Dataset & $\mathcal{S}$ & $\mathcal{R}$ & $\mathcal{Q}$ & $\mathcal{U}$ & $\mathcal{C}$ & $\mathcal{D/N}$ & $\mathcal{T}$ & $\mathcal{O}$ & Dim. \\
     \midrule
     7-Scenes~\cite{shotton2013scene}    & 7  & 26k  & 17k  &        &        &        &      &  & $27$ GB\\
     ScanNet~\cite{dai2017scannet}       & 1.5k & \multicolumn{2}{c}{2.5M} &         &        &        & \cmark & & $1^*$  TB\\
     ScanNet$++$~\cite{yeshwanthliu2023scannetpp} & 460 & \multicolumn{2}{c}{3.7M} &         &        &        & \cmark & & $1.5^*$  TB\\
     CMU Seasons~\cite{Badino2011,Sattler2018CVPR} & 1  & 61k  & 57k & \cmark & \cmark & \cmark &        & \cmark &    $8$ GB  \\
     RobotCar Seasons~\cite{RobotCarDatasetIJRR,Sattler2018CVPR} & 1  & 20k  & 12k  & \cmark & \cmark & \cmark & \cmark & \cmark & $6$ GB \\
     Aachen Day-Night~\cite{Sattler2012BMVC,Sattler2018CVPR} & 1  & 3k   & 922  & \cmark & \cmark &        & \cmark &        &   $2.5$ GB \\
     Cambridge Landmark~\cite{kendall2015posenet} & 5 & 6.8k & 3.4k & \cmark & & & & & $5.6$ GB\\
     KITTI~\cite{geiger2012we} & 1 & 7.5k & 7.5k & \cmark & & & & \cmark & $6.4^*$ GB\\
     Mapillary Metropolis~\cite{metropolis}  & 1  & \multicolumn{2}{c}{27k} & \cmark & & & & \cmark  & $77$ GB\\
     CrowdDriven~\cite{Jafarzadeh_2021_ICCV} & 26 & 1.3k & 1.7k & \cmark & \cmark & \cmark & \cmark & & $2$ GB\\
     \midrule
     Flatlandia & 20 & 6.3k & 2k & \cmark & \cmark & \cmark & \cmark & \cmark & $\mathbf{31}$ \textbf{MB} \\
     \bottomrule
 \end{tabularx}
 \caption{\label{tab:confrontingdatasets} \emph{Popular visual localization datasets}: Comparison between Flatlandia and popular datasets. We report on size (number of geographic areas $\mathcal{S}$, reference images $\mathcal{R}$ and visual queries $\mathcal{Q}$) and on the type of provided data: whether it provides urban images ($\mathcal{U}$), captured by different cameras ($\mathcal{C}$), in both day and night conditions ($\mathcal{D/N}$), at different times/months ($\mathcal{T}$), and if it includes 2D/3D object detections ($\mathcal{O}$). We report also the approximate storage size of the dataset (Dim.). $^*$Estimated from average image size of $400$ kB.}
\end{table}

\subsection{Similar Domain Datasets}~\label{sec:related_dataset}
There are many benchmark datasets for 6DoF localization; 
in Tab.~\ref{tab:confrontingdatasets}, we compare some of the most common ones against the Flatlandia dataset. Some capture \emph{indoors scenes}, like 7-Scenes~\cite{shotton2013scene} 
and ScanNet~\cite{dai2017scannet}; these environments are rich in semantically distinct objects, and can provide 2D/3D detections (ScanNet). However, they generally cover a small area, with frequent overlapping objects when projected onto the 2D map; this makes them unsuitable to train large-scale visual localization methods. Moreover, they are captured with a single camera and in a single session, which might negatively affect generalization capabilities. Conversely, datasets composed only of images acquired from vehicles like KITTI~\cite{geiger2012we}, could introduce biases, as the cameras are bound to the road and can therefore assume a small range of poses in the scene. 
The dataset closest to Flatlandia is CrowdDriven~\cite{Jafarzadeh_2021_ICCV}, which used Mapillary crowd data focusing on day-night settings across different cities; however, it does not provide the 2D/3D object annotations required by our queries. 

\begin{figure*}[t]
        \centering
        \includegraphics[width=\linewidth] {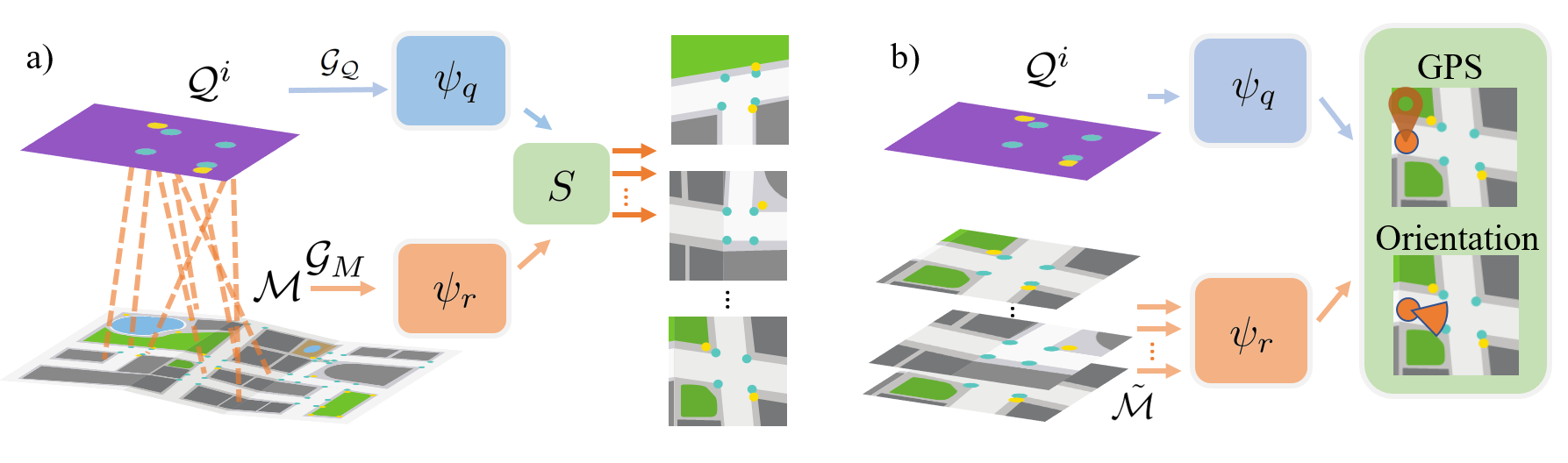}
    \caption{\emph{Concept of the two Flatlandia's tasks}. (Sec.~\ref{sec:problem}) a) Coarse Map Localization: given a query local map, a set of candidate reference nodes (regions) are proposed based on a learned distance of spatial and semantic similarity; b) Fine-grained 3DoF Localization: given a query local map and a set of small regions, the fine-grained 3DoF pose of the camera is regressed.}
    \label{fig:concept}
\end{figure*}

\section{The Flatlandia Problem}\label{sec:problem}
\emph{Object-based 3DoF Visual Localization}, as shown in Fig.~\ref{fig:mainfig}, estimates the location and orientation of a visual query $I$ by comparing two maps (Sec.~\ref{sec:problem_input}): a \emph{local map}, representing the layout of the objects observed in the image; and a \emph{reference map}, representing the known location of objects in the scene. 
We approached this as two tasks: \emph{Coarse Localization} (Sec.~\ref{sec:problem_coarseloc}), to identify a sub-region of the \emph{reference map} compatible with the \emph{local map}; and \emph{Fine-grained Localization} (Sec.~\ref{sec:problem_fineloc}), to combine the \emph{local map} and the retrieved  sub-region and regress the image pose on the reference map. 
%

\subsection{Local and Reference Maps}\label{sec:problem_input}
Flatlandia's maps are defined as a set of objects $j$ with associated a 2D location $c_j \in \mathbb{R}^2$ and a one-hot class encoding $l_j \in \mathbb{N}^{N_C}$, where $N_C$ is the number of possible object classes. 

The \emph{reference map} $\mathcal{M}$ represents the known distribution of objects in the scene; $c_j$ is then defined as the position of object $j$ in a reference system centered in the scene and with y-axis pointing north. In this work, we assume that \emph{i)} any object detected in the query is also localized in $\mathcal{M}$; and \emph{ii)} $c_j$ coincides with the object's center of mass projected on a bird's eye view. These assumptions are suitable for an initial study on object-based 3DoF visual localization, but future works should explore localization with incomplete or noisy reference maps. 

The \emph{local map} $\mathcal{Q}$ represents the distribution of the objects observed in the query; $c_i$ is then the location of object $i$ in a system centered in the camera and with  y-axis parallel to the camera's viewing direction. To account for scale ambiguity, 
the vector of all object locations is normalized. Section~\ref{sec:dataset} provides an in-depth discussion on the maps creation process.

\subsection{Coarse Map Localization}\label{sec:problem_coarseloc}

Given a pair of maps $\mathcal{M}$ and $\mathcal{Q}$, \emph{Coarse Map Localization} retrieves a subset of reference objects $\{c_j\} \in \mathcal{M}$ that exhibit semantic and geometric similarities with $\mathcal{Q}$. This task evaluates the ability to identify within $\mathcal{M}$ a part of the objects of $Q$, which is necessary to infer the pose of the query in the reference map. Additionally, this task provides insight in the possible limitations due to repeated similar patterns at different scales, which is often present in urban environments and could make object-based localization ambiguous.
This tasks, illustrated in Fig.~\ref{fig:concept}a, can be approached as a  
graph-based retrieval problem, as graphs are well suited to model complex relationships between entities. 
Section~\ref{sec:coarseloc} provides an in-depth discussion of the task and the proposed baseline.
%

\subsection{Fine-grained 3DoF Localization}\label{sec:problem_fineloc}

\emph{Fine-grained 3DoF Localization} attempts to regress the 3DoF pose by comparing $\mathcal{Q}$ with the output of \emph{Coarse Map Localization}. This allows to assess the accuracy limitations introduced by abstracting the scene to point-like objects. Given the correct retrieved region and a perfect \emph{local map}, localization could be framed as a 2D alignment problem, mapping $\mathcal{Q}$ onto the matched reference objects. Due to the noise in the local map and errors in the Coarse Map Localization step, we propose instead to learn to directly regress a 3DoF pose.
Section~\ref{sec:fineloc} provides an exhaustive description of the task, and introduces two baseline models; a graph-based baseline, and a baseline composed of fully connected layers.

\section{The Flatlandia Dataset}\label{sec:dataset}
The Flatlandia dataset provides a benchmarking tool for object-based 3DoF visual localization. It is composed of $20$ reference maps $\mathcal{M}$ representing the object distribution in scenes covering $0.006$ to $0.037$ km$^2$ (mean area of $0.025$ km$^2$), and $2$k visual queries $I$ with associated a \emph{local map} $\mathcal{Q}$ and a 3DoF camera pose. 
This provides a set of inputs $[\mathcal{Q}, \mathcal{M}]$ for the localization tasks introduced in Section~\ref{sec:problem}, and a ground-truth 3DoF camera pose to evaluate their output. The whole dataset, as reported in Table~\ref{tab:confrontingdatasets}, amounts to an uncompressed \emph{JSON} dictionary of size $31$~MB; this includes the reference maps ($50$~kB each), two types of local maps ($0.5$~MB each), and additional information such as query images' metadata from Mapillary, ground truth locations, objects' 2D detection.

The dataset is built from crowd-sourced images acquired from the Mapillary platform\footnote{Mapillary: \href{https://mapillary.com}{https://mapillary.com}}. This 
provides a more realistic testing environment, with visual queries captured by different devices (\eg, GoPro's, dash-cams, smartphones), using different types of projections (perspective or fisheye), and in different viewing conditions (\eg in different days and at different times) and from different points of view (\eg from vehicles or by pedestrians). By providing a wide range of viewing directions and type of agents collecting the data, we reduce the risk of bias due to limitations in the possible camera poses, and improve the probability that most objects in the scene are detected in at least a few images. Additionally, Mapillary images come with an estimated GPS pose and orientation, and instance and semantic segmentation.
We here discuss the process used to select and reconstruct the scenes (Section~\ref{sec:scene_reconstruction}), to generate the reference maps (Section~\ref{sec:reference_map}), and to generate the local maps (Section~\ref{sec:local_map}). 

\begin{table}[t]
\scriptsize
\centering
 \begin{tabularx}{\linewidth}{X | c | c c c c c c c} 
 \toprule
City & $|\mathcal{S}|$ & $km^2$ & kB & $\bar{\mathcal{I}}$ & $\bar{\mathcal{O}}$ & $\bar{\mathcal{Q}}$ & $\bar{\mathcal{A}}$ & $\bar{\mathcal{D}}$ \\ [0.5ex]  
\midrule
Vienna & 1 & 0.036 & 11 & 854  & 223 & 420      & 0.070 & 3189 \\
Paris & 7 & 0.027  & 3.9 & 243 & 83 & 83         & 0.047 & 1777 \\
Lisbon & 7 & 0.017 & 6.4 & 328  & 137 & 120      & 0.027 & 5055 \\
Berlin & 2 & 0.024 & 8.7 & 423  & 181 & 89       & 0.039 & 4633 \\
Barcelona & 3 & 0.025 & 3.6 & 211  & 73 & 63     & 0.043 & 1779 \\
\midrule
Flatlandia & 20 & 0.025 & 5.6 & 316  & 118 & 110 & 0.040 & 2967 \\
 \bottomrule
  \end{tabularx}
 \caption{\label{tab:mapo} \emph{Statistics on the Flatlandia dataset.} (Sec.~\ref{sec:dataset}) For each city, we report the number of scenes $\mathcal{S}$ and the average scene size $km^2$; average storage requirement per reference map (kB); number of images used to reconstruct the scene $\bar{\mathcal{I}}$; number of objects detected with a 2D location in a scene $\bar{\mathcal{O}}$; number of local maps in a scene $\bar{\mathcal{Q}}$; area covered by the objects in a scene in km$^2$ $\bar{\mathcal{A}}$; density of objects in a scene expressed in objects per $km^2$ $\bar{\mathcal{D}}$.
 }
\end{table}

\subsection{Scene Reconstruction}\label{sec:scene_reconstruction}

For five cities (Barcelona, Lisbon, Paris, Vienna and Berlin), we acquired all images within $3$~km of the city center, and divided them into overlapping tiles of side $0.5$~km. 
To reduce the risk of reconstruction errors, we generated binary masks occluding any element likely to move (\eg pedestrians and vehicles) or to change appearance (\eg sky, vegetation) and the Mapillary watermark, and we removed nearly duplicated images (\ie consecutive images with low pixel-wise variance). 
The remaining images and their masks were passed to COLMAP~\cite{schonberger2016structure}, to generate a sparse point-cloud model and a 6DoF camera pose for each image. For each scene reconstructed successfully, we removed incorrectly localized cameras (\eg below street level, in the sky), before applying one last iteration of Bundle Adjustment.

\begin{figure}[t]
   \centering
     \includegraphics[trim={0.0 4.5cm 14cm 1cm}, clip, width=\linewidth]{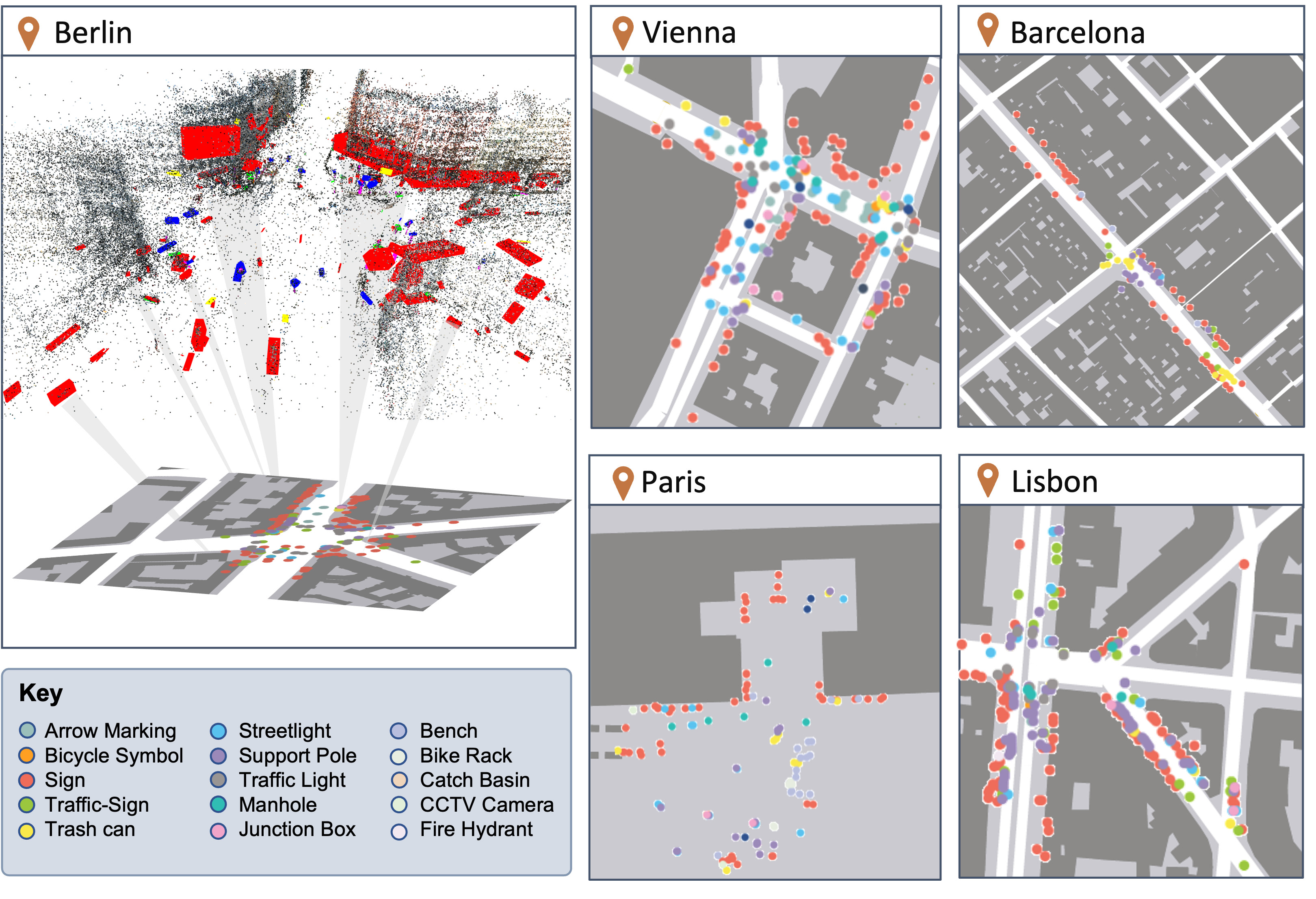}
   \caption{\emph{Example of Reference Map} (Sec:~\ref{sec:reference_map}): SfM reconstruction from Berlin, clustered into 3D objects and projected on a bird's-eye-view map and aligned to GPS coordinates. 
   }
   \label{fig:MAPO}
\end{figure}
 
\subsection{Object Maps Creation}\label{sec:reference_map}
Given the SfM reconstructions, the 3D point cloud can be turned into a set of objects by attributing to each point a class, based on semantic segmentation of the images on which it has matched 2D keypoints, and applying a clustering algorithm. Each cluster can then be reprojected onto the images containing its associated keypoints, and compared against the instance segmentation to correct for possible errors due to proximity of same-class objects or to objects represented by sparse 3D points. Finally, we can define a set of objects by fitting a 3D bounding box to each cluster, and match each object to a set of detections using the 3D point-2D keypoint correspondences and instance segmentation. Table~\ref{tab:mapo} reports the number of scenes reconstructed for each city, the average number of images involved ($316$) and resulting objects ($118$).

The objects and camera poses are then projected on a reference map by \emph{i)} identifying the principal plane of the reconstruction; \emph{ii)} projecting on it the camera poses and the centroids of the objects; \emph{iii)} aligning the projected camera 3DoF poses to the GPS coordinates reported by Mapillary, to find the transformation mapping the reconstruction to GPS coordinates. 
Fig.~\ref{fig:MAPO} provides an example of how a 3D point cloud reconstruction from Berlin is first clustered into 3D objects, and then projected onto a local street map. 

This process, while providing correct object localization and matched detections, is prone to false negatives; scene objects with no corresponding 3D points or instance segmentations without triangulated 2D keypoints are not represented in the reference map. This does not affect the usefulness of the resulting reference maps, but future work could expand the dataset, increasing the number of objects and detections. 

\subsection{Local Maps Creation}\label{sec:local_map}
Given the reference map, we select as visual queries all images that detect at last three of the localized objects; fewer detected objects would make the visual localization problem ill-posed and unsolvable. As reported in Table~\ref{tab:mapo}, only $35\%$ of the Mapillary images ($110$ out of $316$ per scene) satisfy this criterion.

For each visual query, Flatlandia provides two types of local maps. \emph{GT-Based local maps} are obtained by computing the location of the detected  reference maps in the query's reference system. These maps differ from the observed portion of the reference map by a 2D rigid transformation. They have been registered to the reference map by brute-force, to verify all queries can be localized. This involves aligning the local map to all sets of reference objects with the correct class and choosing the set with the lowest residual.

The \emph{Depth-based local maps} provide a more realistic use case, and are obtained by projecting the center of each detection into 3D, before projecting it on the image's ground plane;
the distance between the camera and the detected objects
is estimated as the median over the detections of the 
per-pixel depth estimated with MiDaS~\cite{ranftl2020towards}, a generic monocular depth estimation model pre-trained on a
combination of 12  different datasets. 
The ambiguity inherent to depth estimation and the fact that the center of the detection does not necessarily coincide with the center of the object make the process noisy, with a  
median per-object displacement of $14.8$~m. This allows 
testing the feasibility of object-based visual localization from noisy local maps, and future works  attempting to solve the Flatlandia problem 
are free to choose a different, more accurate, approach for local map estimation.

\begin{figure}[htbp!]
   \centering
    \centering
    \includegraphics[trim={0.5 0cm 16cm 1cm}, clip, width=\linewidth] {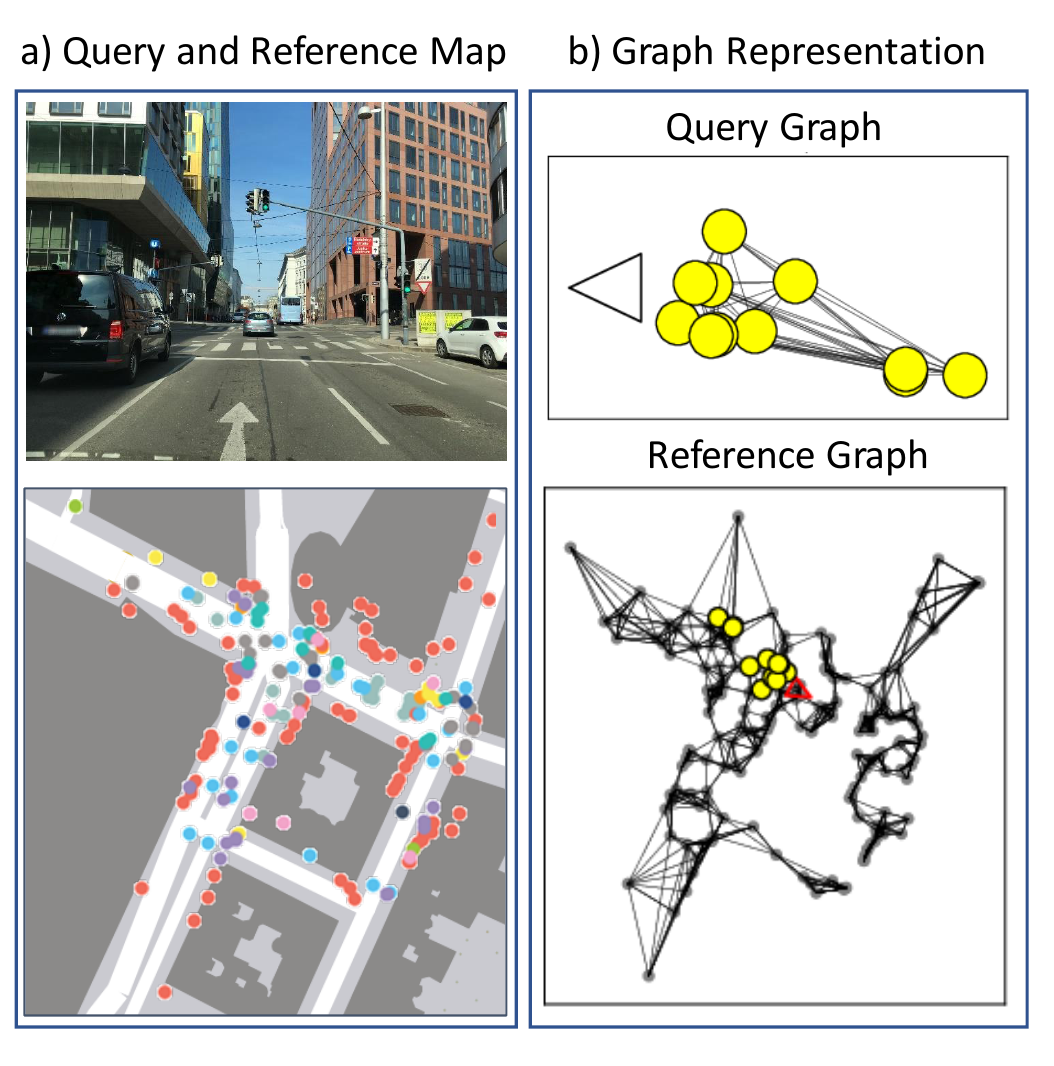}
   \caption{\label{fig:input_to_graphs} \emph{Graph representation of the reference map and the query} (Sec.~\ref{sec:coarseloc}). The visual query and reference map (a) can be represented as object graphs (b).
   The query graph (top) is fully connected; reference graph (bottom) nodes are connected to the seven nearest neighbors. Query nodes (yellow) and the camera pose (red) are overlaid for reference.
   }
\end{figure}

\section{Coarse Map Localization}\label{sec:coarseloc}
As introduced in Section~\ref{sec:problem_coarseloc}, we approach \emph{Coarse Map Localization} as a graph retrieval problem. In Flatlandia's framework, local and reference maps can be naturally described as graphs in which each object $j$ is a node with an initial embedding $\psi_j=[l_j, c_j]$, \ie the concatenation of its class label and its location. The query graph $\mathcal{G}_\mathcal{Q}$ is fully connected, given the small number of nodes. In contrast, in the reference graph $\mathcal{G}_\mathcal{M}$ each node is connected to its $7$ nearest neighbors; this value has been empirically selected to make the reference graph sparse and with a non-trivial topology, while ensuring it has no isolated  sub-graphs, \ie there is at least one path connecting any two nodes. Each edge connecting nodes $i$ and $j$ has an initial embedding $w_{ij} = e^{-\|c_j-c_i\|}$. 
Fig.~\ref{fig:input_to_graphs} exemplifies how the visual query and reference map (\ref{fig:input_to_graphs}a) are turned into graphs (\ref{fig:input_to_graphs}b). 
We train a Graph Attention Network~\cite{velickovic2018graph} (GAT) to process $\mathcal{G}_\mathcal{M}$ and $\mathcal{G}_\mathcal{Q}$, generating node embeddings representative of the local neighborhood around each node. We then define for each reference node $i$ a query similarity score, based on the updated embeddings:
\begin{align}
\label{eq:coarse_sim}
    s_i = \| \psi_i - \psi_\mathcal{Q}\| && \psi_\mathcal{Q} = \langle \psi_j \rangle_{j \in \mathcal{Q}},
\end{align}
\ie the distance in embedding space between the node and an aggregated query embedding $\psi_\mathcal{Q}$ obtained by mean pooling.

These scores are then used to select a \textit{region proposal} $\mathcal{S}$, \ie a set of reference nodes with the neighborhood that most closely resemble the layout observed in the visual query. We consider four possible selection strategies:
\begin{itemize}
    \item $\mathbf{N_{LM}}$: we select the $N$ highest scoring reference nodes, where $N$ is the number of objects detected in the query;
    \item $\mathbf{\mathbf{2N_{LM}}}$: we select the best scoring reference nodes, taking twice as many objects as are detected in the query;
    \item $\mathbf{R_{30}}$: all reference objects within $30$ m of $argmax(s_i)$;
    \item \textbf{GT-BB}:  we select all reference objects within a bounding box fitted to the nodes observed in the query and the GT camera location, and then expanded by $10\%$ of its size; this region proposal does not depend on the predicted scores, but provides a comparison baseline for the other region proposal strategies, and an input to evaluate Fine-grained 3DoF localization independently from this task.
\end{itemize}

\subsection{Training Losses}\label{sec:coarse_graph}
To generate similarity scores (Eq.~\ref{eq:coarse_sim}) representative of the layout similarity between the query and the reference node neighborhood, we train the GAT in supervised fashion using two losses, the Triplet Loss and Neighbor Similarity. 

\textbf{Triplet Loss (Trip)} is a standard approach in retrieval settings~\cite{d2021localized}. Given the embeddings $\psi_i$ and $\psi_\mathcal{Q}$ generated by the GAT, we define as positive matches $M^+$ the nodes of $\mathcal{G}_{\mathcal{M}}$ that represent objects observed in the query. We then define negative matches $M^-$ by randomly sampling the same number of nodes from the nodes of $\mathcal{G}_{\mathcal{M}}$ that do not correspond to detected objects. The model is then trained to minimize the standard triplet loss 
\begin{equation}
\mathcal{L}_t(\mathcal{\psi}_Q, \psi^{+}, \psi^{-} )=\|\psi_{\mathcal{Q}} - \psi^+\|_2 - \|\psi_{\mathcal{Q}} - \psi^-\|_2 + \alpha,    
\end{equation}
 where $\psi^{+/-}=[\psi^{i}_\mathcal{M}]_{i \in M^{+/-}}$ are the embedding from the positive/negative matches, and $\alpha$ is a margin.

\textbf{Neighbor Similarity (Sim)} explicitly exploits the semantic and geometrical similarity between the query and the neighborhood $N(i)$ of the reference node $i$ to define a neighborhood similarity score $S_{i \to \mathcal{Q}}$. Iteratively, all subsets of $N(i)$ are aligned to the query, to identify the subset with the 
smallest alignment residual $d^*_p$.
For that set, the average distance between its nodes and $i$ defines a distance score $d^*_{s}$. The similarity score is then defined as $S_{i \to \mathcal{Q}}=(1-d^*_{p})\cdot(1 - d^*_{s})$. It is maximized if $N(i)$ includes a set of nodes that have the same geographical arrangement of part of the query (first term) and this set includes $i$ or is close to it (second term). The large number of possible subsets makes this process computationally complex, but it is only used in training and $S_{i \to \mathcal{Q}}$ can be pre-computed.
We define the similarity loss as
\begin{equation}
    \mathcal{L}_{d}=\sum_{i}(\mid \| \psi_i - \psi_{\mathcal{Q}} \| - S_{i \to \mathcal{Q}} \mid) \cdot W_{i \to \mathcal{Q}}
\end{equation}
where $W_{i \to \mathcal{Q}} = S_{i \to \mathcal{Q}}\; (\sum_j \|S_{j \to \mathcal{Q}}\|)^{-1}$ are the normalized similarity weights, to maximize the contribution of the highest similarity nodes to the loss. 

(\textbf{Sim + Trip}) combines the two losses, to make Trip more robust against selecting in $M^-$ nodes whose neighborhood is not observed in the image, but has a similar object layout.

\subsection{Metrics}\label{sec:region_selection}
To evaluate the accuracy of Coarse Map Localization approaches, we compare the set $\mathcal{S}$ of nodes selected with the $N_{LM}$, $2N_{LM}$, and $R_{30}$ strategies against $\mathcal{S}^{GT}$, the set of the nodes of $\mathcal{G}_\mathcal{M}$ observed in the visual query. We suggest three metrics to quantify the outcome of this task:

\textbf{Precision (P)}, defined as the fraction of nodes of $\mathcal{S}$ that also belong to $\mathcal{S}^{GT}$, \ie $| \mathcal{S} \cap \mathcal{S}^{GT}| \cdot | \mathcal{S}|^{-1}$.

\textbf{Recall (R)}, defined as the fraction of nodes of $\mathcal{S}^{GT}$ that were retrieved in $\mathcal{S}$, \ie $| \mathcal{S} \cap \mathcal{S}^{GT}| \cdot | \mathcal{S}^{GT}|^{-1}$.

\textbf{Success (S)}, defined as the fraction on visual queries for which $| \mathcal{S} \cap \mathcal{S}^{GT}| \geq 3$, because three correct matches is the minimum necessary (though not sufficient) to unambiguously solve the \emph{fine-grained} localization task.

\section{Fine-grained 3DoF Localization}\label{sec:fineloc}
From the region proposal $\mathcal{S}$, we then obtain a 3DoF camera pose by regressing a position $o \in \mathbb{R}^2$ and an orientation $\theta_0 \in [0, 2\pi]$ in the reference system of $\mathcal{M}$, given initial embeddings $\psi^{U}= \{\psi_i\}_{i \in U}$, where $U \in [\mathcal{M}, \mathcal{S}]$ and $\psi_i = [l_j, c_j]$ is the concatenation of its class label and its location. 

The proposed approach is to train a model $\sigma$ to aggregate information about the semantic and spatial composition of the nodes in $\mathcal{S}$ and $\mathcal{G}_\mathcal{Q}$ in one descriptor vector $\Psi$, which is passed to a \emph{regression module} $\rho$:
\begin{align}
    \label{eq:regression}
    o, \theta_o = \rho(\Psi) && \Psi = \sigma(\{\psi_i\}_{i \in \mathcal{S}}, \{\psi_j\}_{j \in \mathcal{Q}}).
\end{align}

As an initial baseline, we propose a \emph{regression module} composed of three fully-connected (FC) layers with ReLu activation, while for the aggregation module $\sigma(.)$ we consider as possible strategies a graph-based approach, that propagates information alternately between nodes of the same graph and nodes with the same label, and an MLP-based approach, which directly combine the reference and query information using a series of FC layers. 

\subsection{Graph-based Approach}\label{sec:fine_graph}

The objects maps, as discussed in Section~\ref{sec:coarseloc}, can be described as graphs. For this tasks, we propose alternating between aggregating information from nodes associated to objects of the same map, and propagating information between the two maps. First, we define a graph containing all nodes from $\mathcal{G}_{\mathcal{Q}}$ and the nodes from $\mathcal{G}_{\mathcal{M}}$ selected in $\mathcal{S}$. We then draw two types of directional edges, connecting nodes from the same map (\emph{inter-map edges}) or query and reference nodes with the same class label (\emph{intra-map edges}). This defines two graphs, respectively named $\mathcal{G}_{\mathcal{Q} \; \text{xor} \; \mathcal{S}}$ and $\mathcal{G}_{\mathcal{Q} \to \mathcal{S}}$, that share the same node embeddings. When drawing \emph{inter-map edges}, the query nodes are fully connected, while the subgraph belonging to $\mathcal{S}$ connects each node with its 3 nearest neighbors. The initial edge embedding is defined as $\psi_{ij} = [ \|c_j - c_i\|, \theta^{ij}_{\hat n}]$, \ie module and orientation of the segment connecting the objects.
For \emph{intra-map edges} we compute tentative values assuming $\mathcal{Q}$ is centered in $\mathcal{M}$ and facing North.

The aggregation module $\sigma$ is composed of three elements: a projection layer $\Pi(.)$, an intra-map graph network $\sigma_1(.)$, and and inter-map graph network $\sigma_2(.)$. The projection layer uses two FC layers with ReLu activation to project the nodes and edge embeddings into a higher dimensional space:  
\begin{align}\label{eq:projection64}
    \psi'_{ij} = \Pi(\psi_{ij}) && \psi'_{k} = \Pi(\psi_{k}) && \psi'_{ij}, \psi'_{k} \in \mathbb{R}^{64}.
\end{align}
The intra-map graph network $\sigma_1(.)$, composed of two Edge-GAT (EGAT) layers~\cite{101093bbab371} with ReLu activation, is then used to update the node embeddings:
\begin{align}
    \{\psi''_{i}\} = \sigma_1(\{\psi'_i\}, \{\psi'_{ij}\}) && i,ij \in \mathcal{G}_{\mathcal{Q} \; \text{xor} \; \mathcal{S}}, 
\end{align}
while the inter-map graph $\sigma_2(.)$, composed of one EGAT layer with ReLu activation, updates the edge embeddings of $\mathcal{G}_{\mathcal{Q} \to \mathcal{S}}$: 
\begin{align}
    \{\psi''_{ij}\} = \sigma_1(\{\psi''_i\}, \{\psi'_{ij}\}) && i,ij \in \mathcal{G}_{\mathcal{Q} \to \mathcal{S}}. 
\end{align}

This alternated updating scheme is repeated for three times, before aggregating the edge embeddings of $\mathcal{G}_{\mathcal{Q} \to \mathcal{S}}$ in a global embedding $\Psi$ and passing it to the regression module (Eq.~\ref{eq:regression}):
\begin{align}\label{eq:gat}
    \Psi = \text{maxpool}(\{\psi''_{ij}\}) && ij \in \mathcal{G}_{\mathcal{Q} \to \mathcal{S}}.
\end{align}

Drawing inspiration from the encoder blocks used by transformers~\cite{vaswani2017attention}, we also considered including a multi-head attention layer $\tau$ before pooling the edge embeddings:
\begin{align}\label{eq:gat+att}
    \Psi = \text{maxpool}(\tau(\{\psi''_{ij}\})) && ij \in \mathcal{G}_{\mathcal{Q} \to \mathcal{S}}.
\end{align}
We will refer to the two baselines defined by for Eq.~\ref{eq:gat} and Eq.~\ref{eq:gat+att} as \textbf{GAT} and \textbf{GAT+ATT} respectively.

\subsection{MLP-based Approach}\label{sec:fine_fc}
The second baseline approach, \textbf{MLP}, uses FC layers to generate a global embedding for the query $\Psi_Q$, that is combined with the embeddings of the reference objects $j\in \mathcal{S}$ to generate $\Psi$. Recent works~\cite{Guo_2023_WACV} have shown how simple MLP baselines can sometime outperform more sophisticated approaches; we therefore consider the possibility that the limited number of query detections (Sec.~\ref{sec:local_map}) and of the reference nodes in $\mathcal{S}$ could make a simple MLP approach more suitable than the graph-based approach described above. Moreover, the MLP model could implicitly learn to infer the relative poses between objects and whether two objects have the same class from the information encoded in the node embeddings, potentially making the edge information and graph connectivity redundant. In this approach, the projected query embeddings $\psi'_i$ ( Eq.~\ref{eq:projection64}) are aggregated in a vector $\Psi_Q$:
\begin{align}
    \Psi_Q = \text{maxpool}(\psi_i) && i\in \mathcal{Q}.
\end{align}
This is then concatenated to the reference map embeddings, and their information is combined through $\sigma_3$, a module composed of two FC layers with ReLu activation. The global embedding $\Psi$ is then defined as:
\begin{align}\label{eq:mlp}
    \Psi = \text{maxpool}(\sigma_3([\Psi_Q, \psi_j])) &&  j \in \mathcal{S}
\end{align}
As in the previous approach, $\Psi$ is then passed to the regression module (Eq.~\ref{eq:regression}) to generate a 3DoF pose in the reference system of $\mathcal{M}$. We also consider an alternative pipeline with an additional multi-head attention layer $\tau$ before the max-pooling step of Eq.~\ref{eq:mlp}; we refer to this model as \textbf{MLP+ATT}.

\subsection{Evaluation Metrics}\label{sec:3DoF_evaluation}
Given the predicted 3DoF pose $(o,\theta_o)$, we can evaluate the localization accuracy of the proposed baseline approaches by comparing it against the known ground truth 3DoF camera pose $(o^{GT},\theta_o^{GT})$. We separately evaluate localization accuracy with respect to the predicted position ($\Delta o$) and orientation ($\Delta \theta$), which during training define a loss $\mathcal{L}_{3DoF} = \Delta o + \Delta \theta$. 

\textbf{Position accuracy ($\Delta o$)} is the Euclidean distance between the predicted and $GT$ position $\|o-o_{GT}\|_2$. During training, to improve numerical stability, the whole scene (objects and camera's locations) is scaled to fit in a squared area of side $1$; during evaluation, the predicted camera coordinate is re-scaled to GPS coordinates, with error expressed in meters ($m$).

\textbf{Orientation accuracy ($\Delta \theta$)} is $\min(\|\theta_o-\theta_o^{GT}\|, \pi-\|\theta_o-\theta_o^{GT}\|)$, the angular distance between the predicted and $GT$ orientation, expressed in radians in $[0, 2\pi]$. During training, this ensures that both losses operate at similar scales, avoiding the need for a scaling coefficient in $\mathcal{L}_{3DoF}$; at run time, we evaluate in degrees to facilitate interpretation of the results.

\section{Experiments}\label{sec:experiments}

To provide insight on the Flatlandia problem and tasks, we evaluate the models introduced in the previous sections on the Flatlandia dataset. We divide the dataset into the training, validation and testing subsets by randomly splitting the queries of each map in proportion $80\%$, $10\%$ and $10\%$.
Section~\ref{sec:exp_coarse} provides results for the \emph{Coarse Map Localization} task; in Section~\ref{sec:exp_fine} we report performance on \emph{Fine-grained 3DoF Localization} on both synthetic region proposals $\mathcal{S}$ and on the outputs of \emph{Coarse Map Localization}; finally, Section~\ref{sec:exp_soa} compares the best performing 3DoF baseline against popular 3DoF and 6DoF visual localization algorithms.

\begin{table}[b!]
\centering
\scriptsize
 \begin{tabularx}{\linewidth}{l X  c c c c c c } 
 \toprule
 & & \multicolumn{3}{c}{GT-Based local maps } &  \multicolumn{3}{c}{Depth-based local maps} \\
 \cmidrule(lr){3-5}\cmidrule(lr){6-8}
\cmidrule(lr){3-5}\cmidrule(lr){6-8}
M   &\multirow{1}{*}{NS} & P & R &  S & P & R & S \\
\midrule
 & GT-BB & 0.258 & * & * & 0.258 & * & * \\
\midrule
\multirow{3}{*}{Sim}
& $N_{LM}$  & 0.146             & 0.146             & 0.168             & 0.176             & 0.176             & 0.179             \\
& $2N_{LM}$ & 0.133             & 0.266             & 0.328             & 0.149             & 0.299             & 0.326             \\
& $R_{30}$     & 0.105             & 0.439             & \underline{0.533} & 0.134             & \textbf{0.531}    & \underline{0.533} \\

\hdashline
\multirow{3}{*}{Trip}  
& $N_{LM}$  & \underline{0.222} & 0.222             & 0.211             & \underline{0.214} & 0.214             & 0.211             \\
& $2N_{LM}$ & 0.188             & 0.376             & 0.403             & 0.173             & 0.345             & 0.372             \\
& $R_{30}$     & 0.142             & \underline{0.471} & 0.522             & 0.137             & 0.458             & 0.496             \\

\hdashline
 & $N_{LM}$  & \textbf{0.236}    & 0.236            & 0.269            & \textbf{0.219}    & 0.219             & 0.247             \\
Sim + Trip   & $2N_{LM}$ & 0.198             & 0.396            & 0.473             & 0.193             & 0.385             & 0.452             \\
  & $R_{30}$     & 0.148             & \textbf{0.543}   & \textbf{0.600}    & 0.139             & \underline{0.506} & \textbf{0.590}    \\
\bottomrule
\end{tabularx}
     \caption{\label{tab:region_selection_errors}  \emph{Coarse Map Localization Problem} (Sec.~\ref{sec:exp_coarse}): \textbf{P}recision, \textbf{R}ecall and \textbf{S}uccess rate on dataset partitions of the proposed approaches (M). Result shown with different node selection (NS): $N_{LM}$, $2N_{LM}$, $R_{30}$ (* indicates 100\% by design. Best values in bold, second best underlined).}
\end{table}
\subsection{Coarse Map Localization}\label{sec:exp_coarse}

To evaluate Coarse Map localization, we provide numerical results for models trained using the loss functions and selection strategies introduced in Section~\ref{sec:coarse_graph} 
and the input local maps generated in Section~\ref{sec:local_map}. 
 The models were trained for $800$ epochs using the Adam optimizer and a batch size of $20$. Training on Flatlandia dataset required less than $7$ hours.

For each model, Table~\ref{tab:region_selection_errors} reports the average precision and recall, and the overall success rate. To provide a comparison baseline, we include precision and recall for the GT-BB region proposals (Sec,~\ref{sec:coarse_graph}). 
The task presents a difficult challenge for all proposed models, with the best average recall and success scores $0.543$ and $0.600$ respectively. 
Precision scores are similarly low, with a maximum of $0.236$, which is however close to the comparison baseline of GT-BB ($0.258$). The combined loss (Sim+Trip) leads to the best performance,
while Sim leads to the overall worst results. 
In terms of selection strategies, $N_{LM}$ achieves the highest precision and lowest recall, while $R_{30}$ has the lowest precision and highest recall; $2N_{LM}$ provides a compromise between the two, with more balanced results. Finally, the performance on GT-based and Depth-based local maps are comparable; as expected, using the GT-based local maps results in better performance.

\begin{figure}[thbp!]
   \centering
    \centering
    \includegraphics[trim={0 0cm 15cm 0}, clip, width=\linewidth] {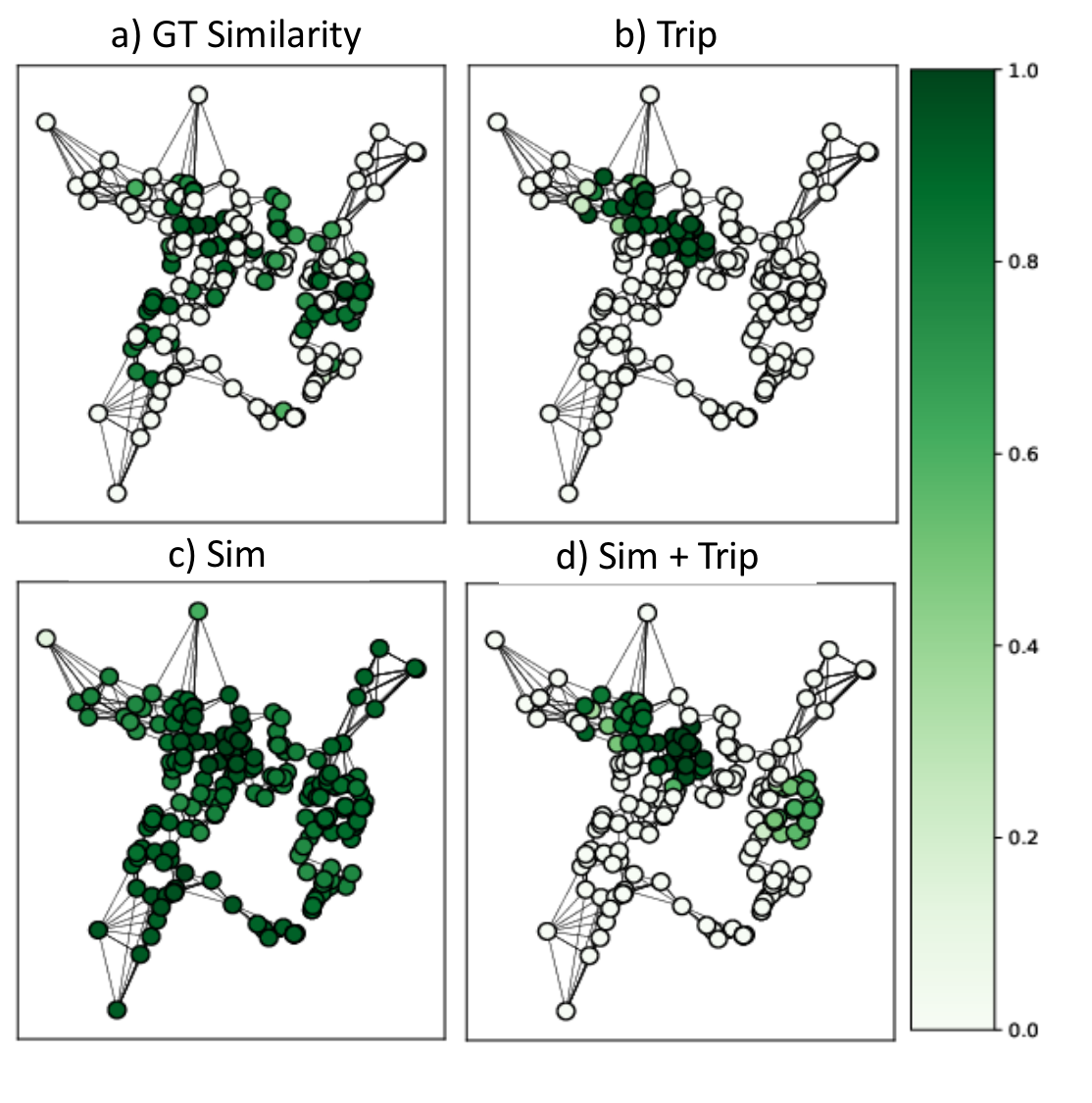}
   \caption{\label{fig:graph_to_node_similairty} \emph{Similarity scores between the reference nodes and the query.} (Sec.~\ref{sec:exp_coarse}). GT similarity scores computed by brute-force alignment (a) vs. values predicted by the Trip (b), Sim (c) and Sim+Trip (d) models for the graphs of Fig.~\ref{fig:input_to_graphs}.
   }
\includegraphics[trim={0cm 1cm 17cm 0cm}, clip, width=\linewidth] {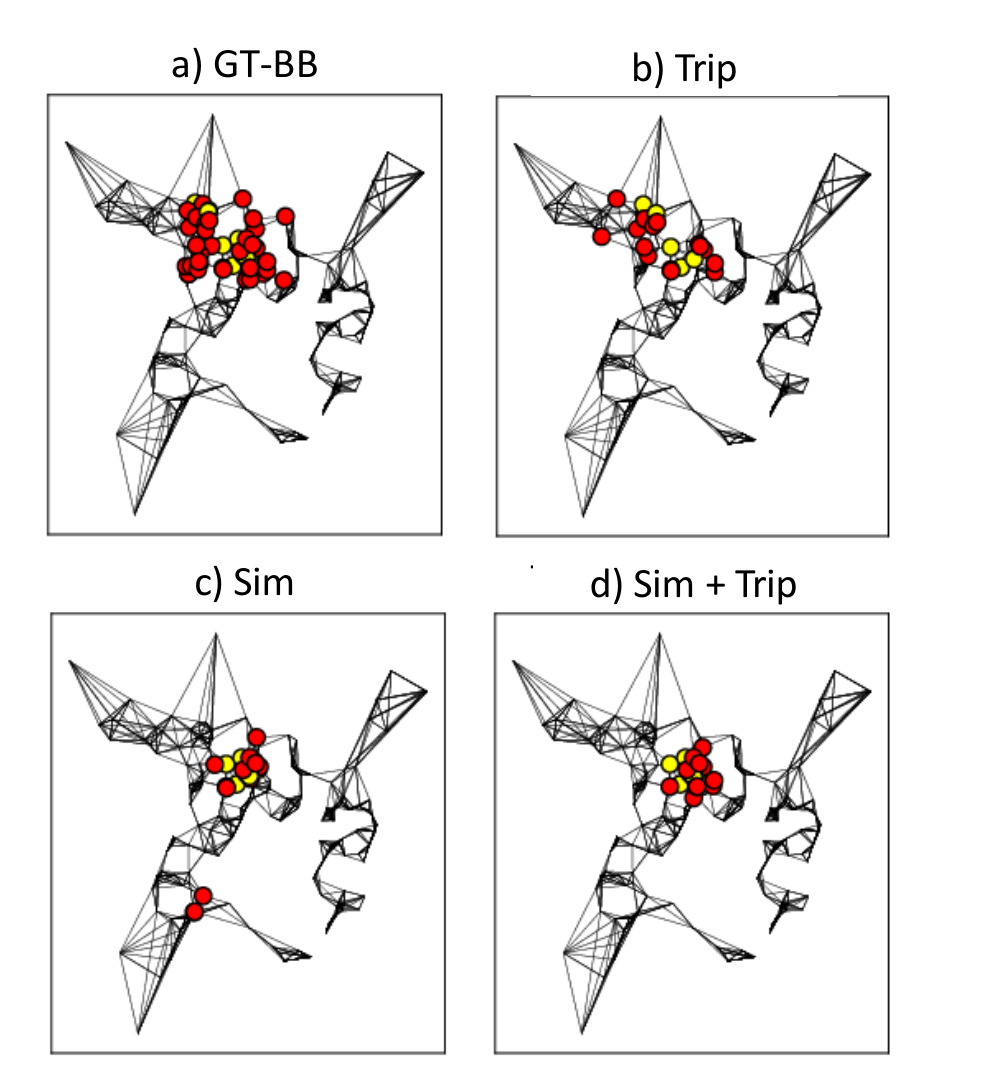}
   \caption{\label{fig:map_localization_output}\emph{Coarse Map Localization} (Sec.~\ref{sec:exp_coarse}). Region proposals $\mathcal{S}$ extracted from predicted similarity scores (Fig.~\ref{fig:graph_to_node_similairty}b,c,d), 
   compared against the \textbf{GT-BB} (a) region proposal (Sec.~\ref{sec:coarseloc}). True positives are highlighted in yellow, false positives in red. 
   }
   
\end{figure}

To provide more insight on these results, Fig.~\ref{fig:graph_to_node_similairty} and Fig.~\ref{fig:map_localization_output} exemplify the similarity scores and region proposals obtained from the 
graphs of Fig.~\ref{fig:input_to_graphs}. Looking at the predicted similarity scores in Fig.~\ref{fig:graph_to_node_similairty}, all models correctly attribute high scores to nodes close to the query; however, all models, especially Sim, generate a large number of false positives, \ie nodes with a high similarity score but far from the query. The large density of objects with high scores can explain the low precision and recall, especially for the $N_{LM}$ and $2N_{LM}$ selection strategies. 
Looking at the region proposals generated the $2N_{LM}$ selection strategy (Fig.~\ref{fig:map_localization_output}) all models retrieve some of the correct nodes (yellow) but also several false positives (red), which are not part of the query but are close to it. This suggests that a more sophisticated model, or a better selection strategy, can achieve better recall performance. In contrast, the high density of objects explains the low precision of GT-BB.


\subsection{Fine-grained 3DoF Localization}\label{sec:exp_fine}

\begin{table*}[]
\centering
\begin{tabularx}{\linewidth}{l c@{\hskip 0.1 in} c@{\hskip 0.1 in} c@{\hskip 0.1 in} c c c c | c@{\hskip 0.1 in} c@{\hskip 0.1 in} c@{\hskip 0.1 in} c c c c}
\toprule
 &  \multicolumn{7}{c}{GT-Based local maps} & \multicolumn{7}{c}{Depth-based local maps} \\  
\cmidrule(lr){2 - 8} \cmidrule(lr){9 - 15}  
&  \multicolumn{4}{c}{$\% $ below(m / $ ^ \circ$)} &  \multicolumn{2}{c}{$\mu \pm \sigma $} & &  \multicolumn {4}{c}{$\% $ below(m / $ ^ \circ$)} &  \multicolumn{2}{c}{$\mu \pm \sigma $} &  \\
\cmidrule(lr){2 - 5} \cmidrule(lr){6 - 7} \cmidrule(lr){9 - 12} \cmidrule(lr){13 - 14} 
Model & 0.5/2 & 1/5 & 5/10 & 10/20 &  $\mu$(m) & $\mu$($^\circ$) & QT(s) & 0.5/2 & 1/5 & 5/10 & 10/20 &  $\mu$(m) & $\mu$($^\circ$) & QT(s) \\
\midrule
MLP & 0.0 & 0.5 & 10.0 & 37.0 & \textbf{12.5} $\mathbf{\pm}$ \textbf{17.3} & \textbf{13.0} $\mathbf{\pm}$ \textbf{36.0} & 0.02 & 0.0 & 1.9 & 16.6 & 36.0 & \textbf{13.3} $\mathbf{\pm}$ \textbf{16.6} & \textbf{12.9} $\mathbf{\pm}$ \textbf{38.5} & 0.02 \\
MLP w.Att. & 0.0 & 0.0 & 13.3 & 32.2 & \underline{12.8 $\pm$ 16.9} & 14.2 $\pm$ 36.7 & 0.02 & 0.0 & 0.5 & 5.7 & 25.6 & \underline{15.5 $\pm$ 15.9} & 17.9 $\pm$ 36.3 & 0.02 \\
GAT & 0.0 & 0.0 & 3.3 & 13.7 & 19.0 $\pm$ 18.3 & 22.1 $\pm$ 35.1 & 0.03 & 0.0 & 0.0 & 4.7 & 17.1 & 21.3 $\pm$ 21.4 & \underline{15.9 $\pm$ 37.0} & 0.03 \\
GAT w.Att. & 0.0 & 0.0 & 8.1 & 21.3 & 18.2 $\pm$ 19.4 & \underline{14.2 $\pm$ 34.5} & 0.02 & 0.0 & 0.5 & 2.4 & 16.1 & 19.3 $\pm$ 18.8 & 21.1 $\pm$ 31.4 & 0.02 \\
\bottomrule
\end{tabularx}
\caption{\label{tab:GT_fineloc} \emph{Fine-grained Localization from GT-BB region proposals} (Sec.~\ref{sec:exp_fine}): we report the median ($\mu$) position ($m$) and orientation ($^\circ$) error for models trained using the GT or Depth-based local maps. QT(s) is the mean query time (s).}
\end{table*}

\begin{table*}[t]
\begin{tabularx}{\linewidth}{l c@{\hskip 0.1 in} c@{\hskip 0.1 in} | c@{\hskip 0.1 in} c@{\hskip 0.1 in} c@{\hskip 0.1 in} c@{\hskip 0.1 in} c c c | c@{\hskip 0.1 in} c@{\hskip 0.1 in} c@{\hskip 0.1 in} c@{\hskip 0.1 in} c c c }
\toprule
 & & & \multicolumn{7}{c}{GT-based local maps} & \multicolumn{7}{c}{Depth-based local maps} \\  
\cmidrule(lr){4 - 10} \cmidrule(lr){11 - 17}
&  \multicolumn {2}{c}{RP} & \multicolumn {4}{c}{$\% $ below(m / $ ^ \circ$)} &  \multicolumn{2}{c}{$\mu \pm \sigma $} &  & \multicolumn {4}{c}{$\% $ below(m / $ ^ \circ$)} &  \multicolumn{2}{c}{$\mu \pm \sigma $} &  \\
\cmidrule(lr){2 - 3} \cmidrule(lr){4 - 7} \cmidrule(lr){8 - 9} \cmidrule(lr){11 - 14} \cmidrule(lr){15 - 16}
Model & S & T & 0.5/2 & 1/5 & 5/10 & 10/20 &  $\mu$(m) & $\mu$($^\circ$) & F $\%$ & 0.5/2 & 1/5 & 5/10 & 10/20 &  $\mu$(m) & $\mu$($^\circ$) & F $\%$\\
\midrule
MLP & \cmark &  & 0.0 & 0.0 & 1.9 & 9.0 & \textbf{32.9} $\mathbf{\pm}$ \textbf{35.1} & \textbf{44.1} $\pm$ \textbf{49.6} & 13 & 0.0 & 0.0 & 2.8 & 10.0 & \textbf{35.4} $\pm$ \textbf{33.3} & 41.6 $\pm$ 50.5 & 12 \\
MLP w.Att. & \cmark & & 0.0 & 0.0 & 2.8 & 8.5 & 37.9 $\pm$ 37.2 & 49.7 $\pm$ 54.8 & 13 & 0.0 & 0.5 & 1.9 & 8.1 & 35.7 $\pm$ 38.5 & \textbf{38.7} $\pm$ \textbf{52.5} & 12 \\
GAT & \cmark &  & 0.0 & 0.0 & 0.0 & 1.4 & \underline{35.6 $\pm$ 33.9} & 51.2 $\pm$ 50.2 & 13 & 0.0 & 0.0 & 1.4 & 2.8 & \underline{35.6 $\pm$ 33.5} & \underline{39.4 $\pm$ 51.1} & 12 \\
GAT w.Att. & \cmark &  & 0.0 & 0.0 & 1.4 & 5.2 & 37.1 $\pm$ 36.0 & \underline{48.4 $\pm$ 54.4} & 13 & 0.0 & 0.0 & 0.5 & 4.7 & 38.2 $\pm$ 35.7 & 40.3 $\pm$ 52.3 & 12 \\
\hdashline
MLP & & \cmark & 0.0 & 0.0 & 2.8 & 12.3 & \textbf{31.8} $\pm$ \textbf{31.5} & \textbf{35.6} $\pm$ \textbf{51.4} & 7 & 0.0 & 0.0 & 4.7 & 7.6 & \textbf{31.6} $\pm$ \textbf{35.9} & \textbf{34.1} $\pm$ \textbf{53.2} & 7 \\
MLP w.Att. & & \cmark &  0.0 & 0.0 & 4.3 & 10.4 & 36.9 $\pm$ 35.4 & \underline{36.6 $\pm$ 51.6} & 7 & 0.0 & 0.5 & 2.8 & 10.0 & 35.8 $\pm$ 34.9 & \underline{37.0 $\pm$ 52.1} & 7 \\
GAT & & \cmark & 0.0 & 0.0 & 0.0 & 3.3 & \underline{33.7 $\pm$ 29.4} & 37.5 $\pm$ 46.1 & 7 & 0.0 & 0.0 & 2.4 & 4.7 & \underline{34.7 $\pm$ 30.5} & 41.2 $\pm$ 49.4 & 7 \\
GAT w.Att. & & \cmark & 0.0 & 0.0 & 0.9 & 5.2 & 38.9 $\pm$ 31.8 & 37.1 $\pm$ 47.7 & 7 & 0.0 & 0.0 & 0.5 & 3.8 & 37.7 $\pm$ 32.3 & 41.9 $\pm$ 50.1 & 7 \\
\hdashline
MLP & \cmark & \cmark & 0.0 & 0.0 & 6.2 & 12.8 & \textbf{25.1} $\pm$ \textbf{32.1} & \underline{25.8 $\pm$ 48.9} & 7 & 0.0 & 0.5 & 6.2 & 12.8 & \textbf{25.3} $\pm$ \textbf{35.4} & \textbf{24.4} $\pm$ \textbf{52.0} & 9 \\
MLP w.Att. & \cmark & \cmark &0.0 & 0.9 & 4.7 & 11.4 & \underline{28.1 $\pm$ 35.5} & \textbf{22.5} $\pm$ \textbf{52.0} & 7 & 0.0 & 0.5 & 3.8 & 8.5 & \underline{30.2 $\pm$ 34.8} & \underline{28.4 $\pm$ 48.7} & 9 \\
GAT & \cmark & \cmark & 0.0 & 0.0 & 0.9 & 5.7 & 31.4 $\pm$ 32.3 & 32.1 $\pm$ 48.2 & 7 & 0.0 & 0.0 & 2.4 & 5.7 & 32.2 $\pm$ 31.9 & 34.8 $\pm$ 50.9 & 9 \\
GAT w.Att. & \cmark & \cmark & 0.0 & 0.0 & 0.9 & 6.6 & 30.7 $\pm$ 31.7 & 29.0 $\pm$ 47.5 & 7 & 0.0 & 0.5 & 0.9 & 6.6 & 32.7 $\pm$ 31.8 & 32.4 $\pm$ 49.5 & 9 \\
\bottomrule
\end{tabularx}
\caption{\label{tab:fineloc_from_coarse_gt} \emph{Fine-grained localization from Coarse Localization output} (Sec.~\ref{sec:exp_fine}): performance on GT and Depth-based queries using as reference the region proposals analyzed in Tab.\ref{tab:region_selection_errors}. $F \%$ is the fraction of queries for which localization failed.}
\end{table*}

To evaluate Fine-grained 3DoF localization, we train the models of Section~\ref{sec:fineloc} to regress 3DoF camera pose from the region proposals generated by Coarse Map Localization. 
In addition to the median location and orientation error of the regressed poses, we report the fraction of queries outperforming various accuracy thresholds, following the convention of~\cite{Jafarzadeh_2021_ICCV}. These thresholds are $0.5$, $1$, $5$, $10$~m and $2$, $5$, $10$, $20^\circ$ for localization and orientation performance respectively.  All baseline models were trained on one GPU and 2 CPU cores for 1000 epochs. 
This required up to 24 hours. To speed up training and improve generalization, the data is augmented by applying rigid roto-translations to the local maps. 

\begin{figure}[t]
        \centering
        \includegraphics[width=\linewidth] {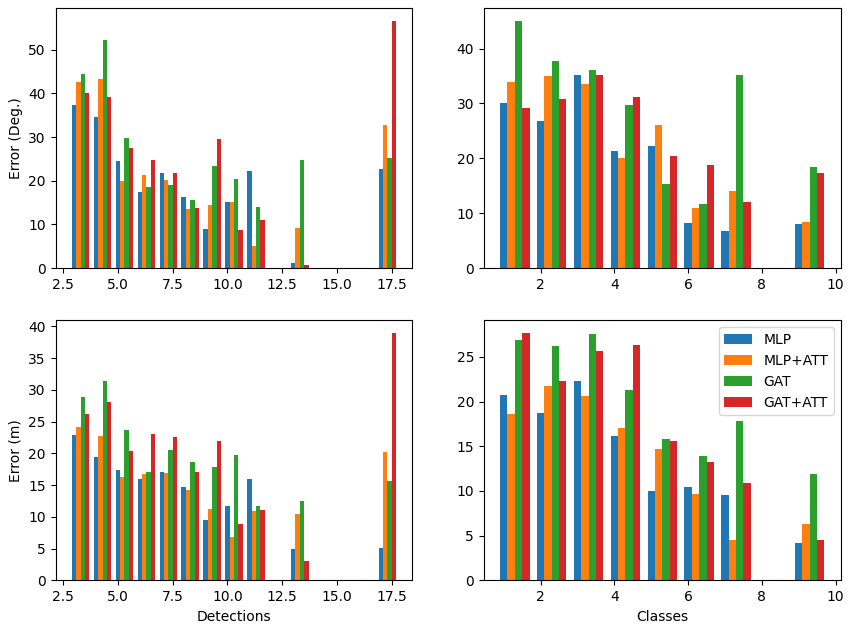}
    \caption{\emph{Distribution of orientation and localization error} (Sec: \ref{sec:exp_fine}): angular and orientation errors as a function of the query's number of classes and detections. There is a positive correlation between an increase in the number/class variability of detections and the reliability of the reconstructed poses.}
    \label{fig:cluster}
\end{figure}

First, evaluating the models on the GT-BB region proposals allows
to study Fine-grained Localization independently from the Coarse Map Localization task. The results, reported in Table~\ref{tab:GT_fineloc}, show how the MLP baseline largely outperforms the GAT and Attention-based baselines. This is reflected by the position error, $12.5$~m and $19$~m for the MLP and GAT model respectively, and by the orientation error, respectively $13^\circ$ and $22.1^\circ$. This suggests that 
the \emph{region proposals} provided as input to 3DoF localization are too small and near fully connected, making graphs an inefficient way to aggregate the information and likely resulting in over-smoothing, as discussed in Section~\ref{sec:fineloc}. Conversely, the positive results provided by the MLP model suggest that the object maps used by Flatlandia contain sufficient information to make 3DoF pose regression possible. A $12.5$~m localization error is worse than the average GPS localization accuracy ($4.9$~m\footnote{\href{https://www.gps.gov/systems/gps/performance/accuracy/}{www.gps.gov/performance/accuracy}}), but the proposed models constitute an initial baseline, and are not meant to be a solution to the Flatlandia problem. To explore the link between localization accuracy and number/variety of detections in the query, Fig.~\ref{fig:cluster} presents the per-query position and orientation accuracy in function of the number of detections and class labels.  
The resulting distribution suggests that the accuracy is proportional to number and variety of objects in the query. We point out that this holds especially for the \emph{classes}, while the \emph{detection} plots show more outliers, especially for the models that are noisier on this task (GAT).

Then, Table~\ref{tab:fineloc_from_coarse_gt} reports the performance obtained using as input 
the region proposals generated by Coarse Map Localization 
using the $2N_{LM}$ selection method. 
The noisy region proposals make the problem harder, leading to errors twice as large as those of Tab.~\ref{tab:GT_fineloc}; moreover, for an average $10\%$ of the queries, the proposed region did not contain enough objects of the appropriate classes, making the problem unsolvable. These limitations are due to the low recall of the Coarse Map Localization models, which results in only $24\%$ or less of the query objects being included in the region proposal; conversely, GT-BB has perfect recall by construction. Using depth or GT-based local maps has less impact on the performance, leading to an average performance gap of $1.2$~m. This is small, compared with the large average noise on the depth-based object locations (Sec.~\ref{sec:local_map}), suggesting that the proposed baselines can cope with a significant level of noise in the local maps. Finally, the MLP baseline still outperforms the other approaches, with on average an $11\%$ performance gain over MLP with attention, a $15 \%$ gain over GAT, and a $17 \%$ performance gain with respect to GAT with Attention.

\subsection{Flatlandia vs. other Localization problems}\label{sec:exp_soa}
\begin{table}[t]
\scriptsize
\begin{tabularx}{\linewidth}{l@{\hskip 0.1 in} c@{\hskip 0.1 in} c@{\hskip 0.1 in} c@{\hskip 0.1 in} c@{\hskip 0.1 in} c@{\hskip 0.1 in} c@{\hskip 0.05 in} c@{\hskip 0.1 in} c}
\toprule
&  & \multicolumn {4}{c}{$\% $ below(m / $ ^ \circ$)} &  \multicolumn{2}{c}{$\mu \pm \sigma $} & \\
 \cmidrule(lr){3 - 6} \cmidrule(lr){7 - 8}
Model & MB &  0.5/2 & 1/5 & 5/10 & 10/20 &  $\mu$(m) & $\mu$($^\circ$) & QT(s) \\
\midrule
VLAD~\cite{arandjelovic2013all} & 390 & 5.2 & 6.2 & 40.3 & 73.5 & \underline{4.8 $\pm$ 5.4} & 6.5 $\pm$ 15.0 & 0.06 \\
HLoc~\cite{sarlin2019coarse} & 449  & 1.4 & 9.0 & 40.3 & 64.0 & 5.5 $\pm$ 11.1 & \underline{4.6 $\pm$ 13.2} & 1.95 \\
PoseNet~\cite{kendall2015posenet}& 150 & 0.9 & 9.0 & 51.2 & 71.1 & \textbf{3.6 $\pm$ 5.1} & \textbf{2.2 $\pm$ 21.3} & \textbf{0.01} \\
\midrule
{\textit{MLP w.Depth}} & \textbf{2} & 0.0 & 1.9 & 16.6 & 36.0 & 13.3 $\pm$ 16.6 & 12.9 $\pm$ 38.5 &\underline{0.02} \\
\bottomrule
\end{tabularx}
\caption{\label{tab:6DoF} \emph{Evaluation on the Flatlandia dataset of popular 6DoF methods} (Sec.~\ref{sec:exp_soa}): we compared with the MLP baseline on depth-based local maps from Table~\ref{tab:GT_fineloc} popular 6DoF visual localization methods. We also report the average size of the scene's model (MB), 
and the query time (QT).}
\end{table}

The baseline models can then be compared against existing methods. 
Table~\ref{tab:6DoF} reports the performance of popular 6DoF localization methods on the Flatnaldia dataset, compared against the best scoring baseline model (MLP on depth-based local maps and GT-BB region proposals). This comparison covers various methods: standard image retrieval (VLAD~\cite{arandjelovic2013all}), retrieval + pose refinement via PnP (HLoc~\cite{sarlin2019coarse}) and implicit scene model (PoseNet~\cite{kendall2015posenet}).
The 6DoF models perform better than the 3DoF baseline, with VLAD, HLoc and PoseNet improving on its accuracy by $1.4$, $1.6$ and $3.8$ times respectively. Conversely, the 3DoF baseline's reference data is more storage efficient, requiring on average $2$~MB per scene
to store maps, detections and initial embeddings
. This is less than $1\%$ of the memory requirement for reference images and point clouds needed by VLAD and HLoc ($390$ and $449$~MB respectively).
The 3DoF baseline is also computationally efficient, with query time comparable with PoseNet ($10^{-2}$ seconds), and better than retrieval ($10^{-1}$ seconds) and HLoc ($2$ seconds).

\begin{table}[t]
\scriptsize
\begin{tabularx}{\linewidth}{ l@{\hskip 0.1 in} c@{\hskip 0.1 in} c@{\hskip 0.1 in} c@{\hskip 0.1 in} c@{\hskip 0.1 in} c@{\hskip 0.1 in} c@{\hskip 0.1 in} c@{\hskip 0.1 in}} 
\toprule
& & \multicolumn {4}{c}{$\% $ below(m / $ ^ \circ$)} &  \multicolumn{2}{c}{$\mu \pm \sigma $} \\
\cmidrule(lr){3 - 6} \cmidrule(lr){7 - 8}
Model & L &0.5/2 & 1/5 & 5/10 & 10/20 &  $\mu$(m) & $\mu$($^\circ$) \\ 
\midrule
OrienterNet~\cite{sarlin2023orienternet}  & 128 & 0.0 & 0.0 & 4.7 & 16.6 & 32.7 $\pm$ 41.2 & 20.5 $\pm$ 60.9\\
OrienterNet~\cite{sarlin2023orienternet} pin.  & 128 & 0.0 & 0.0 & 8.0 & 22.0 & 21.7 $\pm$ 37.3 & 12.0 $\pm$ 58.6 \\
\midrule
\textit{MLP w.Depth.} & 158 & 0.0 & 1.9 & 16.6 & 36.0 & 13.3 $\pm$ 16.6 & 12.9 $\pm$ 38.5\\
\textit{MLP w.Depth.} pin & 158 & 0.0 & 1.9 & 10.0 & 25.1 & 14.3 $\pm$ 17.4 & 12.6 $\pm$ 39.6\\
\midrule
\end{tabularx}
\caption{\label{tab:OrNet} \emph{Evaluation of OrienterNet~\cite{sarlin2023orienternet} on Flatlandia}(Sec.~\ref{sec:exp_soa}). OrienterNet performance on the Flatlandia queries, with as reference OSM tiles of size L centered in a point within $5$m of the GT GPS pose. Results are compared against the Mapillary refined GPS poses. In pin, only images captured with pinhole cameras are used.} 
\end{table}

Table~\ref{tab:OrNet} provides a comparison against OrienterNet~\cite{sarlin2023orienternet}, a popular 3DoF approach. Originally, \cite{sarlin2023orienternet} was trained on pinhole images and using as ground truth the GPS labels generated by Mapillary; for fairness, this evaluation also uses Mapillary data, and reports both performance on the full Flatlandia test set, and on the subset of images acquired by pinhole cameras ($76\%$ of the queries). Using only pinhole images, the performances of the MLP baseline and~\cite{sarlin2023orienternet} are comparable.
When using images with radial distortion, \cite{sarlin2023orienternet} accuracy worsens, going from $21.7$~m to $32.7$~m due to the model not generalizing to different camera models. In contrast, Flatlandia's baseline performance goes from $14.3$~m to $13.3$~m.

\section{Conclusion}
Flatlandia is a novel approach to 3DoF visual localization, designed to develop efficient and privacy-preserving models that localize using the layout of common objects. 

Extensive evaluation of seven baseline models shows that the proposed tasks can be solved, and provide initial comparison baselines for the evaluation of future, more sophisticated approaches. Specifically, the initial baselines (Sec.~\ref{sec:exp_coarse}) show how a simple graph model can partially retrieve the region of the reference map observed in the query, with a recall score of $0.543$. Further evaluation (Sec.~\ref{sec:exp_fine}) shows that these noisy region proposals are enough to localize the query with a median accuracy of $30$~m. Sec.~\ref{sec:exp_fine} then shows how improving the region proposal (recall~$1$, precision $0.26$) can reduce the localization error to $< 15$~m, comparable with the performance of state-of-the-art 3DoF methods on scenes of the size of our maps using a single visual query (Sec.~\ref{sec:exp_soa}). This suggests that dedicated methods could significantly improve 3DoF localization accuracy from a single image. Moreover, these early results confirm that object-only localization is possible, despite abstracting the scene's and query content to point-sized objects on 2D maps and despite the possible presence of repeated object patterns in urban scenes.

Comparison against 6DoF localization models (Sec.~\ref{sec:exp_soa}) then showed how the Flatlandia framework is efficient, requiring $100\times$ less storage memory than traditional approaches to store the reference model, and with query time compatible with PoseNet. We are, therefore, confident that it will provide a helpful benchmark for the development of future methods.

\subsection{Limitations}
The Flatlandia problem requires scenes with varied and abundant objects, resulting in queries with multiple object detections and a rich reference map. This framework is, therefore, not suitable for environments with few objects. Moreover, errors in the reference map such as mislabeling or missing objects could result in incorrect localization; this could limit scalability, since in principle services like OSM or public surveying can provide this information, but their quality and availability can significantly change depending on the location.



\subsection{Future Work}
We expect future work based on the Flatlandia problem will include more sophisticated approaches to visual localization, exploiting the recent advancements in graph-based object matching or diffusion to explicitly establish matches between the object detections and the reference map annotation. Concerning the dataset, we expect future improvements will focus on the main limitations of Flatlandia, such as increasing the number of objects (see the limitations above) or including noise/omissions in reference maps. 
Addressing these dataset challenges would be beneficial not only for 3DoF but 6DoF localization, as they are a currently active area of research.  

\section*{Acknowledgment}
This project has received funding from the European Union's Horizon 2020 research and innovation programme under grant agreement No 870743, and
the project Future Artificial Intelligence Research (FAIR) – PNRR MUR Cod. PE0000013 - CUP: E63C22001940006.

\ifCLASSOPTIONcaptionsoff
  \newpage
\fi

\bibliographystyle{IEEEtran}
\bibliography{flatlandia,datasets}

\begin{IEEEbiography}[{\includegraphics[width=1in,height=1.25in,clip,keepaspectratio]{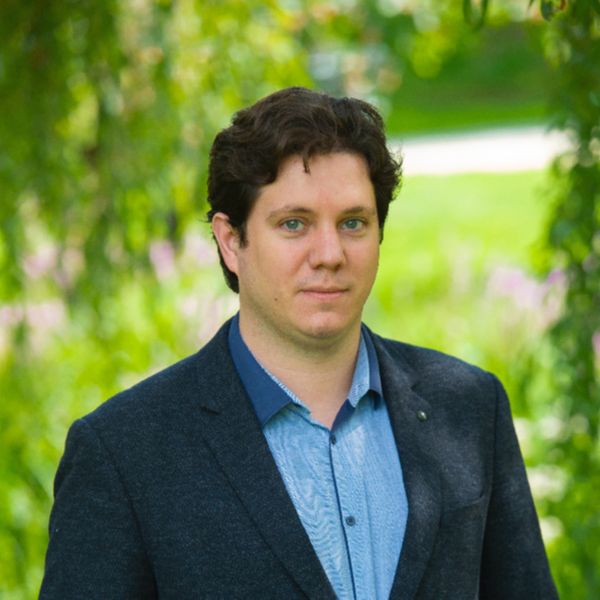}}]{Matteo Toso}
is a PostDoc researcher it the PAVIS research line of the Italian Institute of Technology (IIT), working on object-based visual localization. He holds a PhD degree in Electronic Engineering from the University of Surrey (United Kingdom), with a thesis on Continuous Learning for Inverse Problems with applications to Structure from Motion; and a MSc in Condensed Matter Physics from the University of Trieste (Italy).
\end{IEEEbiography}

\begin{IEEEbiography}[{\includegraphics[width=1in,height=1.25in,clip,keepaspectratio]{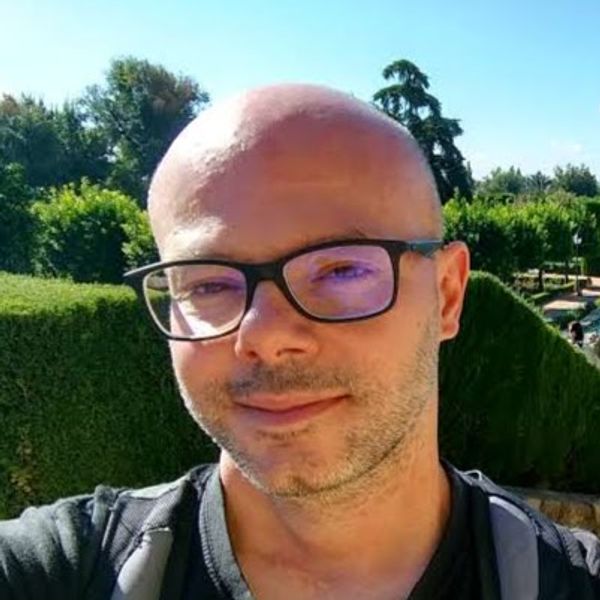}}]{Matteo Taiana} is a postdoctoral researcher at the Pattern Analysis and computer VISion (PAVIS) Research Line of the Italian Institute of Technology (IIT). He holds a MSc degree in Computer Engineering from Milan Polytechnic (Italy) and a PhD degree from Instituto Superior Tecnico (Portugal), with a thesis on Pedestrian Detection. His interests include 3D Computer Vision, Tracking and Machine Learning.\end{IEEEbiography}

\begin{IEEEbiography}[{\includegraphics[width=1in,height=1.25in,clip,keepaspectratio]{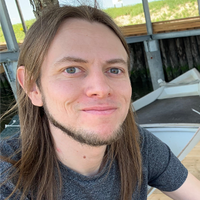}}]{Stuart James}
Researcher (Assistant Professor) in Computer Vision at the Istituto Italiano di Tecnologia (IIT). Stuart's research focus is on Visual Reasoning to understand the layout of visual content from Iconography (e.g., Sketches) to 3D Scene understanding and their implications on methods of interaction. He was involved in the coordination and implementation of the MEMEX EU H2020 project for increasing social inclusion with Cultural Heritage. Stuart has previously held PostDoc positions at IIT, University College London (UCL), and the University of Surrey. Also, at the University of Surrey, Stuart was awarded his PhD. Stuart continues to hold an honorary position at UCL and UCL Digital Humanities and collaborates with researchers across universities and institutes, reviewing and organizing workshops.
\end{IEEEbiography}

\begin{IEEEbiography}[{\includegraphics[width=1in,height=1.25in,clip,keepaspectratio]{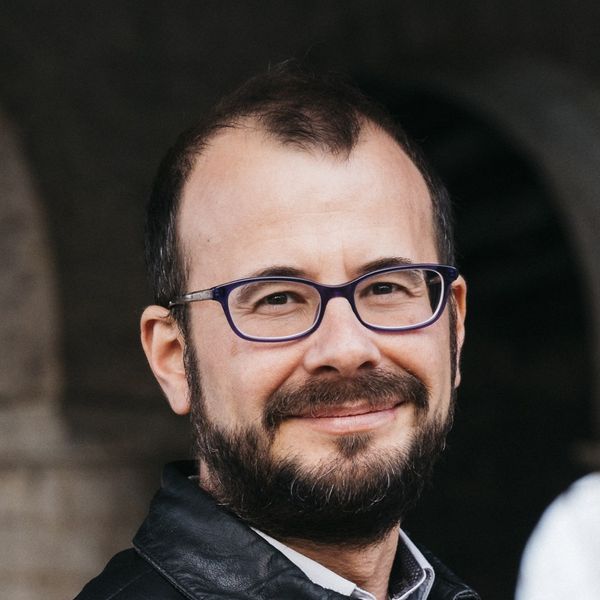}}]{Alessio Del Bue}
Alessio Del Bue is a tenured senior researcher leading the PAVIS (Pattern Analysis and computer VISion) research line of the Italian Institute of Technology (IIT) in Genova, Italy. Previously, he was a researcher in the Institute for Systems and Robotics at the Instituto Superior Técnico (IST) in Lisbon, Portugal. Before that, he obtained my Ph.D. under the supervision of Dr. Lourdes Agapito in the Department of Computer Science at Queen Mary University of London.His current research interests are related to 3D scene understanding from multi-modal input (images, depth, audio) to support the development of assistive Artificial Intelligence systems. He is co-author of more than 100 scientific publications, in refereed journals and international conferences, member of the technical committees of important computer vision conferences (CVPR, ICCV, ECCV, BMVC, etc.), and he serves as an associate editor of Pattern Recognition and Computer Vision and Image Understanding journals. Finally, Dr. Del Bue is an IEEE and ELLIS member in the recently formed Genoa unit.
\end{IEEEbiography}

\end{document}